\documentclass[10pt]{cai}

\begin{document}
\def\conferenceyear{2025}
\volumeheader{38}{0}
\begin{center}

\title{Fast Graph Neural Network for Image Classification}
\maketitle

\thispagestyle{empty}

\begin{tabular}{cc}
Mustafa Mohammadi Gharasuie\upstairs{\affilone,*}, Luis Rueda\upstairs{\affilone}
\\[0.25ex]
{\small \upstairs{\affilone} School of Computer Science, University of Windsor, ON, CA} \\
\end{tabular}
  
\emails{
  \upstairs{*}Mohamm6m@uwindsor.ca 
}
\vspace*{0.2in}
\end{center}

\begin{abstract}
The rapid progress in image classification has been largely driven by the adoption of Graph Convolutional Networks (GCNs), which offer a robust framework for handling complex data structures. This study introduces a novel approach that integrates GCNs with Voronoi diagrams to enhance image classification by leveraging their ability to effectively model relational data. Unlike conventional convolutional neural networks (CNNs), our method represents images as graphs, where pixels or regions function as vertices. These graphs are then refined using corresponding Delaunay triangulations, optimizing their representation. The proposed model achieves significant improvements in both preprocessing efficiency and classification accuracy across various benchmark datasets, surpassing state-of-the-art approaches, particularly in challenging scenarios involving intricate scenes and fine-grained categories. Experimental results, validated through cross-validation, underscore the effectiveness of combining GCNs with Voronoi diagrams for advancing image classification. This research not only presents a novel perspective on image classification but also expands the potential applications of graph-based learning paradigms in computer vision and unstructured data analysis.
\end{abstract}

\copyrightnotice


\section{Introduction}
The field of image classification has undergone a paradigm shift with the emergence of deep learning techniques, particularly, Graph Convolutional Networks (GCNs), which have transformed the way complex image-related tasks are approached. Due to their inherent ability to process data in a graph-based format, GCNs are especially well-suited for applications where relational context and structural information play a crucial role. This introduction explores recent advancements in image classification using GCNs, with a particular focus on the integration of superpixels and wavelet techniques.

Graph Convolutional Networks (GCNs) have become a powerful framework for image classification, largely due to their ability to capture and process the non-Euclidean structure of data. Zhou et al. provided a comprehensive review of GCN applications across various domains, highlighting their effectiveness in image classification tasks \cite{zhou2020graph}. Their study emphasized that, unlike traditional Convolutional Neural Networks (CNNs), GCNs excel at capturing long-range dependencies and complex relational patterns within images.
Meanwhile, wavelet techniques \cite{Wavelet_vasudevan2023image} have transformed image processing by introducing a multi-resolution analysis framework that is essential for applications such as image compression, feature extraction, and noise reduction.

In this context, Wavelet transforms have been utilized in conjunction with GCNs for feature extraction in image classification, denoising, compression and other tasks. Wavelets provide a multi-resolution image analysis, capturing spatial and frequency domain information \cite{mallat1989theory}. 

This paper presents an innovative framework that integrates Graph Convolutional Networks (GCNs) with Voronoi diagrams for image classification, leveraging their exceptional ability to model relational data. Unlike traditional Convolutional Neural Networks (CNNs), our approach represents images as graphs using superpixels, where relational contexts are captured as edges.
The incorporation of Voronoi diagrams enables the partitioning of images into distinct, yet interconnected regions, facilitating a more context-aware graph construction. This method enhances the understanding of spatial and contextual relationships within the image data. Additionally, we propose a novel graph construction technique that effectively captures both local and global image features, improving the model’s ability to recognize intricate patterns and subtle variations.
The primary contributions of this paper are as follows:
(i) transforming images into graphs using superpixels without requiring preprocessing;
(ii) introducing a novel image representation based on Voronoi diagrams, where superpixels define the regions; and
(iii) developing a dual-graph representation model for images using Delaunay triangulations.
 
\section{Related Work}
Simple Linear Iterative Clustering (SLIC), introduced in \cite{8-SLIC-Achanta2012SLIC}, is a method designed to generate superpixels by grouping pixels based on two key criteria: color similarity and spatial proximity. The core of the algorithm lies in minimizing a distance measure, \( D \), which combines the color distance \( d_{lab} \) in the CIELAB color space with the spatial distance \( d_{xy} \) in the image coordinate space. The composite distance measure used by SLIC to form superpixels is given by:

\begin{equation}
    D = \sqrt{(d_{lab})^2 + \left(\frac{d_{xy}}{S}\right)^2 \cdot m^2}     \, ,
\label{eq:SLIC_formula}
\end{equation}
Here, \( S \) represents the spacing between superpixel centers, while \( m \) is a compactness parameter that balances color similarity and spatial proximity. The algorithm iteratively refines superpixel boundaries by recalculating the distance \( D \) and reassigning pixels to the nearest cluster center until convergence.

A variant, Manifold SLIC \cite{8-SLIC-Achanta2012SLIC}, extends the SLIC algorithm to handle data on a manifold rather than in Euclidean space. In image processing, a manifold can be seen as a curved surface in a high-dimensional space, where its intrinsic geometry better captures the underlying structure of image data.

The manifold distance \( D_m \) in Manifold SLIC is generally expressed as:

\begin{equation}
    D_m = \sqrt{(d_{lab})^2 + \left(\frac{d_{geo}}{S}\right)^2 \cdot m^2} \, ,    
\label{eq:ManifoldSLIC_formula}
\end{equation}

In this formula, \( d_{lab} \) represents the color distance in the CIELAB space, while \( d_{geo} \) denotes the geodesic distance in the manifold, measuring the shortest path along its surface. \( S \) is the superpixel grid interval, and \( m \) controls the balance between color similarity and spatial proximity, akin to the compactness parameter in SLIC.

An improvement in superpixel segmentation is the \emph{Superpixels None-Iterative Clustering} (SNIC) algorithm \cite{15-SNIC-achanta2017superpixels}, which builds upon SLIC while enhancing efficiency. Unlike SLIC, which initializes and iteratively refines clusters, SNIC begins by treating each pixel as a cluster and progressively merges them based on color and spatial proximity. This method ensures superpixels adhere closely to object boundaries while maintaining computational simplicity.

SNIC improves SLIC by addressing connectivity and speed limitations. Using a priority queue for superpixel growth, it ensures connected regions while reducing computational cost. Achieving comparable or superior segmentation quality with faster execution \cite{15-SNIC-achanta2017superpixels}, SNIC is a valuable tool in computer vision.

TurboPixels \cite{9-turbopixel} is another superpixel generation technique designed to create compact, uniform regions that closely follow image boundaries. It grows contours from seed points iteratively to cover the image, relying on geodesic distance to align superpixel boundaries with contours. The growth process is governed by a partial differential equation (PDE) based on the level set method, formulated to minimize an energy functional \( E \) that incorporates intensity gradients and contour curvature. While the full formulation involves calculus of variations, it can be expressed as:

\begin{equation}
    E(C) = \int_{C} g(I(x, y)) \cdot \lVert \nabla C(x, y) \rVert + \alpha \cdot \kappa(C(x, y)) \, ds
\label{eq:TurboPixel_formula}
\end{equation}

In this formula, \( C \) represents the contour, \( g(I(x, y)) \) is a stopping function derived from the image intensity \( I \) that slows contour evolution near strong edges, \( \nabla C \) is the contour gradient, and \( \kappa(C(x, y)) \) denotes its curvature. The parameter \( \alpha \) controls the trade-off between adherence to image boundaries and contour smoothness, while integration is performed over the contour length.

Another approach, Linear Spectral Clustering (LSC) \cite{11-LinearSpectralClustering7814265}, is an advanced image segmentation technique that utilizes spectral clustering methods. It efficiently segments an image by mapping pixels into a lower-dimensional space, making clusters more distinguishable.

The core of LSC involves constructing a similarity matrix \( W \) and computing the eigenvectors of the Laplacian matrix \( L = D - W \), where \( D \) is the diagonal degree matrix of the vertices. The main formula for identifying clusters within the transformed space is derived from the eigenvalue problem:

\begin{equation}   
    L_\mathbf{u} = \lambda D_\mathbf{u} \, .
\label{eq:LSC_formula}
\end{equation}
Here, \(\mathbf{u}\) represents the eigenvector corresponding to the eigenvalue \(\lambda\), and solving this problem yields a set of eigenvectors that are then used for clustering. The smallest non-zero eigenvalues and their corresponding eigenvectors capture the most significant relationships between pixels, which are indicative of the underlying segments within the image.

\begin{figure}
    \centering
    \includegraphics[width=0.7\textwidth]{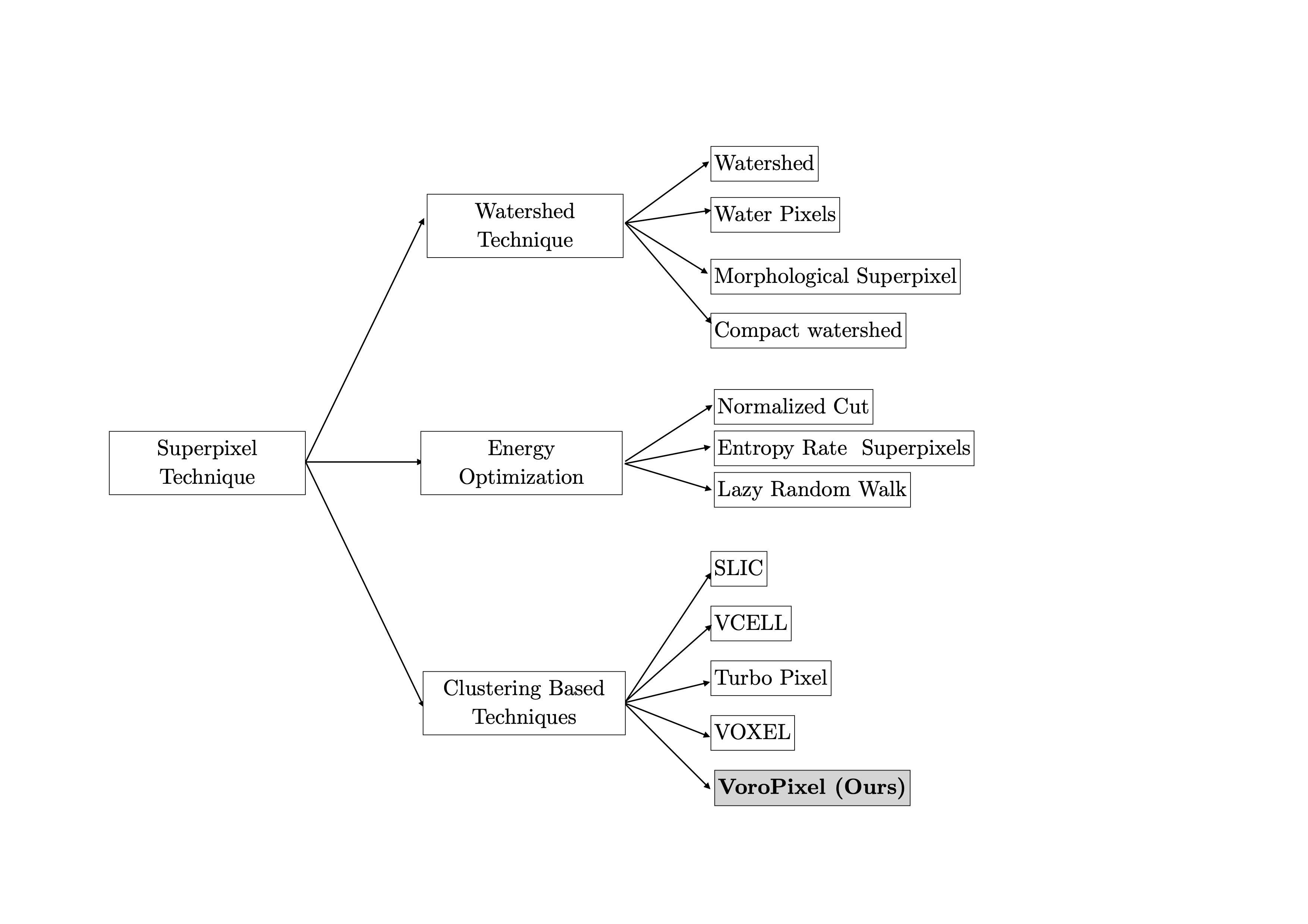}
    \caption{Hierarchical view of different superpixel techniques for images.}
    \label{fig:variation_superpixel}
\end{figure}

Integrating Graph Convolutional Networks (GCNs) with superpixels is a crucial step in image classification, enabling efficient graph construction, enhanced feature extraction, and improved classification accuracy. Recent studies have demonstrated the effectiveness of this approach in processing high-resolution images and complex visual patterns.
While various methods exist for generating superpixels, most require preprocessing the image for projection, adding significant computational overhead to the machine learning downstream task. Kumar et al. in \cite{6-kumar2023extensive} provide a categorized review of superpixel techniques. Fig. \ref{fig:variation_superpixel} illustrates different approaches in this domain.

In addition, the role of Voronoi diagrams in image classification has taken a significant step in this direction. The Voronoi diagram's capability to create distinct regions based on proximity offers a unique way to segment images, which can complement the graph-based approach of GCNs. Aurenhammer et al. provided a comprehensive review of Voronoi diagrams and their properties \cite{4-aurenhammer1991voronoi}. In the context of image classification, Voronoi diagrams can be used to refine the graph representation of an image further, where each cell in the diagram represents a vertex, enhancing the GCN’s ability to process and classify complex image data. However, Utilizing Voronoi diagrams is not limited to just the classification. for example, on stereo satellite images, \cite{voronoiSateliteImages} proposes a “continuous–discrete–continuous” cyclic LSM method, based on the Voronoi diagram. Or, Tac-VGNN \cite{tacvgnn} introduces a Voronoi-based GNN approach to tactile servoing, leveraging spatially structured tactile signals for precise pose estimation and control. This method demonstrates the potential of GNNs in encoding complex tactile information. Other works related to the Voronoi Diagrams and Delaunay triangulation can be found in \cite{adaptiveCentroidalVoronoi}, 

\section{The Proposed Method: VGCN}\label{proposed_method} 
To introduce the main idea of VGCN, we first present relevant preliminaries, notation, and formalisms.

Let 
$\textbf{V}=\{\mathbf{p_1}, \dots,\mathbf{p_m}\}\subset \mathbb{R}^2$ where $m\ge 2 $ and $\mathbf{x_i}\neq \mathbf{x_j}$ for $i\neq j, i,j\in I_n=\{1,\dots,n \}$.
We call a region given by 
\begin{equation}
    \mathbf{V(p_i)} = \{\mathbf{x} \big| \|\mathbf{x-x_i}\|\leq \|\mathbf{x-x_j}\| for j\neq i, j\in I_n\}    
    \label{eq:planar_voronoi_polygon}
\end{equation}
the planar ordinary Voronoi Polygon associated with $\mathbf{p_i}$ (or the Voronoi polygon of $\mathbf{p_i}$), and the set given by
\begin{equation}
    \mathcal{V}=\{\mathbf{V(p_1)},\dots,\mathbf{V(p_m)}\}
    \label{eq:planar_ordinary_voronoi_diagram}
\end{equation}
the planar ordinary Voronoi diagram generated by $\mathbf{P}$ (or Voronoi diagram of $\mathbf{P}$).
In equation \ref{eq:planar_voronoi_polygon}, $I_n$, $\mathbf{x}$, and $\mathbf{x_i}$ refers to the space, coordinate of a point in the space, and the $generator_i$ respectively.
We call $\mathbf{p_i}$ of $\mathbf{V(p_i)}$ the \textit{generator point}, \textit{generator}, or \textit{centroids} of the $i$th Voronoi polygon, and the set $\mathbf{P}=\{\mathbf{p_1},\dots, \mathbf{p_m}\}$ the \textit{generator set} of the Voronoi diagram $\mathcal{V}$. In some works, the generator points are referred to as a \textit{site}.

\begin{figure}
    \centering
    \includegraphics[width=0.3\textwidth]{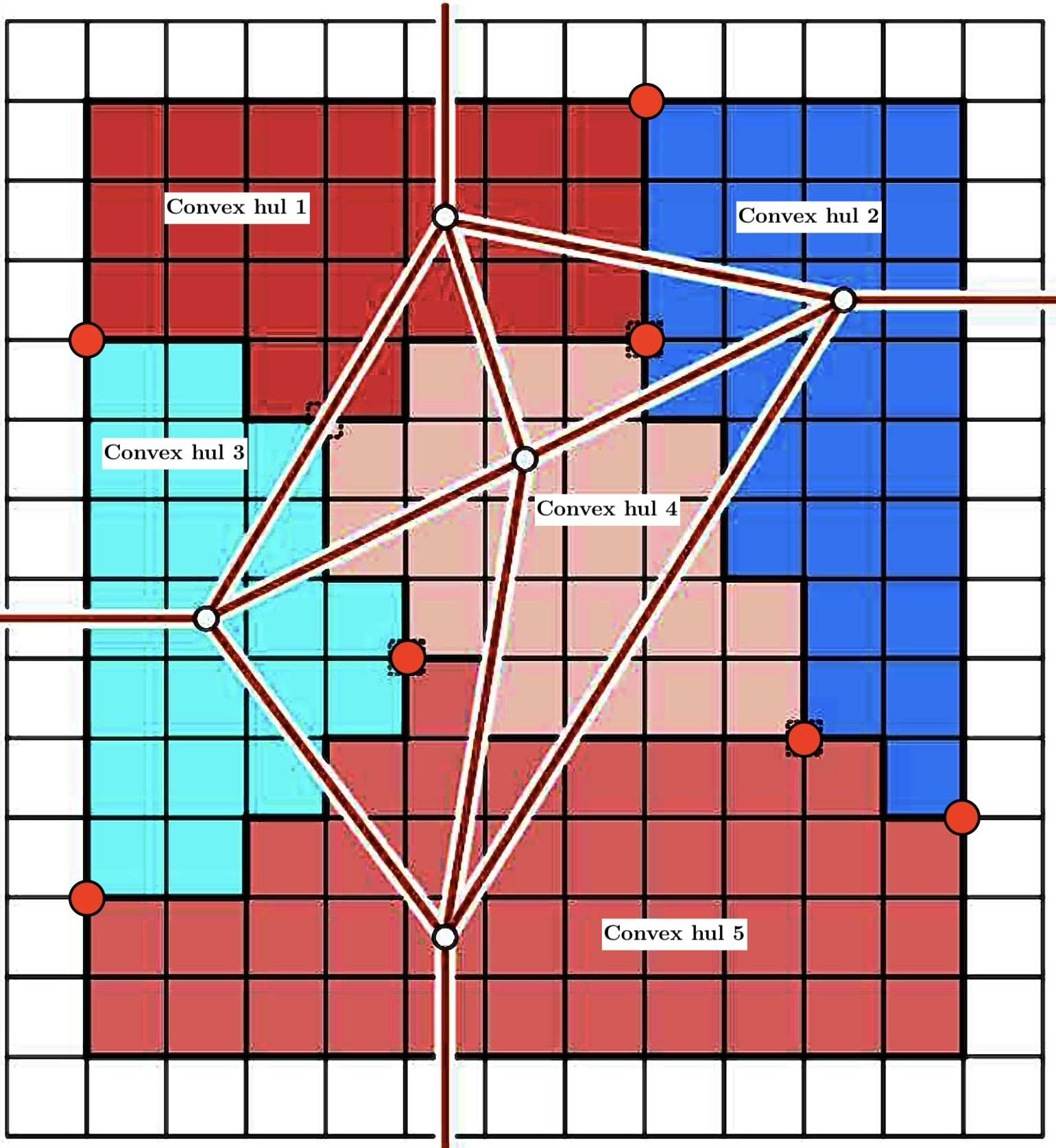}
    \caption{Schematic view of a digital Delaunay triangulation superimposed on the digital Voronoi diagram obtained by flooding.}
    \label{fig:Digitized_vor_tess}
\end{figure}

If $\mathbf{V(p_i)}\cap \mathbf{V(p_j)}\neq \emptyset$, the set $\mathbf{V(p_i)} \cap \mathbf{V(p_j)}$ gives a Voronoi edge. We use $e(\mathbf{p_i}, \mathbf{p_j})$ or $e_{ij}$ instead of $\mathbf{V(p_i)} \cap \mathbf{V(p_j)}$. if $e_{ij}$ is neither empty nor a point, we say that the Voronoi polygons $\mathbf{V(p_i)}$ and $\mathbf{V(p_j)}$ are adjacent.
We can simply change the formula to fit it to the image. The result is a \textit{planar digitized Voronoi diagram} defined as follows:
\begin{equation}
    \textbf{Im}\mathcal{V}=\{\mathbf{\textbf{Im}(V(p_1))},\dots, \mathbf{\textbf{Im}(V(p_m))}\} \, ,
\label{eq:Planar_digitized_voronoi_diagram}
\end{equation}
which \textbf{Im}$\mathbf{V(p_i)}$ is a \textit{digitized Voronoi region}, Im$\delta V(p_i)$ the border of a digitized Voronoi region \textbf{Im}$\mathbf{V(p_i)}$\cite{12-digitized_vor_tess}. It is straightforward to see that in the Digitalized Voronoi diagram $1\le m \le n$. Fig. \ref{fig:Digitized_vor_tess}  shows the digitized Voronoi diagram in the background of the graph.

The proposed method utilizes Voronoi Diagrams and Delaunay Triangulation graphs in the GCN. We name this method Voronoi Graph Convolution Network (VGCN). 

Graph Convolutional Networks (GCNs), introduced in \cite{GNNModelScarselli2009TheGN}, are deep learning models designed for inference on graph-structured data. They capture graph dependencies through message passing between vertices. The core idea is to learn an embedding for each vertex \( v \), incorporating both its attributes and the influence of its neighbors. The vertex representation is iteratively updated using the following equation:

\begin{equation}
\begin{aligned}
    h_v^{(k)} &= \text{UPDATE}^{(k)} \left( h_v^{(k-1)}, \right. \\
    &\quad \left. \text{AGGREGATE}^{(k)} \left( \{ h_u^{(k-1)} : u \in \mathcal{N}(v) \} \right) \right)
\end{aligned}
\label{eq:GCNFormula}
\end{equation}
\noindent where \( h_v^{(k)} \) represents the feature vector of vertex \( v \) at iteration \( k \), \( \mathcal{N}(v) \) denotes its neighbor set, and UPDATE and AGGREGATE are differentiable functions for feature updating and aggregation.

Message Passing in GCNs enables vertices to exchange information with neighbors \cite{18-MessagePassing-Gilmer2017NeuralMP}, allowing vertex representations to reflect both features and graph structure. At each iteration, a vertex sends a message based on its current state, which its neighbors aggregate to update their own state. The general message-passing rule is given by:

\begin{equation}
b_{v}^{(k)} = \sum_{u \in \mathcal{N}(v)} B^{(k)} \left( h_v^{(k-1)}, h_u^{(k-1)}, e_{uv} \right)
\label{eq:MessagePassingFormula}
\end{equation}

\begin{equation}
h_v^{(k)} = U^{(k)} \left( h_v^{(k-1)}, m_{v}^{(k)} \right)
\end{equation}

\noindent where \( b_{v}^{(k)} \) is the aggregated message received by vertex \( v \) at iteration \( k \), \( B^{(k)} \) and \( U^{(k)} \) are the message and update functions, respectively, and \( e_{uv} \) denotes the edge attributes between vertices \( u \) and \( v \).

Graph Attention Networks (GATs) enhance GCNs by incorporating an attention mechanism into the aggregation function \cite{19-GAT-Velickovic2018GraphAN}, allowing vertices to dynamically weigh the importance of their neighbors' messages. This enables selective focus on more relevant neighbors. The attention coefficient between two vertices \( u \) and \( v \) in GATs is computed as follows:

\begin{equation}
\alpha_{uv} = \frac{\exp\left( \text{LeakyReLU} \left( \mathbf{a}^T [W h_u || W h_v] \right) \right)}{\sum_{k \in \mathcal{N}(v)} \exp\left( \text{LeakyReLU} \left( \mathbf{a}^T [W h_u || W h_k] \right) \right)} \, ,
\label{eq:GAT}
\end{equation}

\begin{figure}
    \centering
    \includegraphics[width=0.5\textwidth]{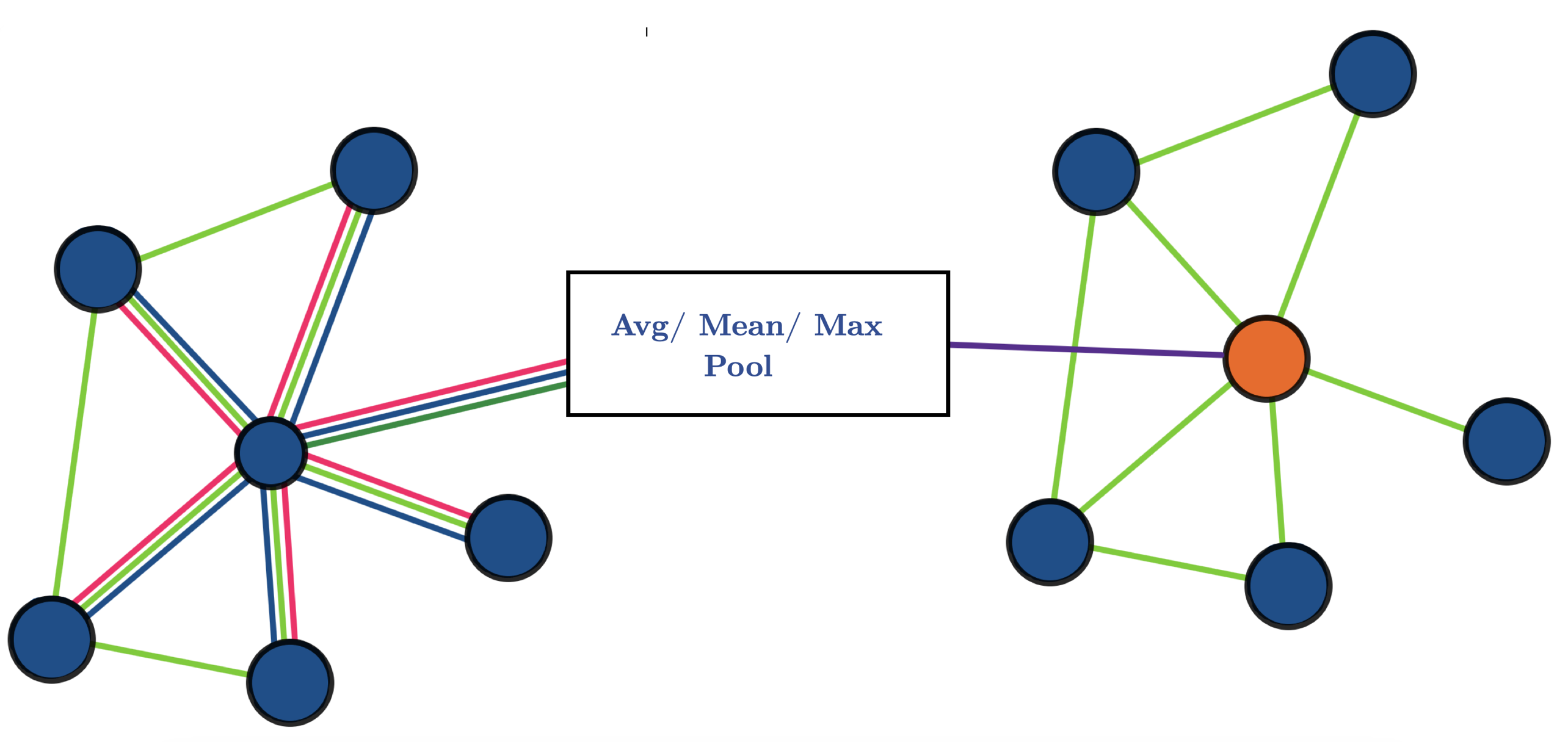}
    \caption{Graph Attention Mechanism. The figure shows how a vertex will be outweighed by another vertex in the attention mechanism. Also, The figure shows the different types of aggregations. Different colors mean different attention in the model. }
    \label{fig:GAT}
\end{figure}

The vertex representation is then updated by weighting the neighbor's features by the attention coefficients:

\begin{equation}
h_v^{\prime} = \sigma \left( \sum_{u \in \mathcal{N}(v)} \alpha_{uv} W h_u \right) \, ,
\end{equation}

\noindent where \( \mathbf{a} \) is a learnable parameter vector, \( W \) is a weight matrix applied to each vertex, \( || \) denotes concatenation, and \( \sigma \) is a non-linear activation function. This attention mechanism allows GATs to dynamically learn the relative importance of neighbors in updating a vertex's feature representation.

Fig. \ref{fig:GAT} illustrates the attention and aggregation mechanism concerning the \( \alpha \) parameter.

\section{The VGCN Architecture}
The first step in the pipeline is to convert the input image into the corresponding Delaunay triangulation graph.  
The first step in the pipeline is to transform the input image into its corresponding Delaunay triangulation graph.  
Algorithm \emph{Delaunay Triangulation} provides the pseudocode for this step in detail.

\begin{algorithm}
\caption{Delaunay Triangulation (Algorithm III-C)} 
  \textbf{Input:} image, $\mathbf{Img}$, number of expected Delaunay triangulation graph's vertices, $m$\\
  \textbf{Output:} Delaunay triangulation graph, $G_{Tess}$\\
    \begin{algorithmic}[1]
    \State{L, S, P are label map, segments and the centroids respectively}    
    \State{$\mathbf{L},\mathbf{S} , \mathbf{P} \gets \text{SNIC($\mathbf{Img}$, $m$)}$}
    \State{$\textbf{X}, \mathbf{E} \gets 
    \text{
    Douglas\_Peucker($\mathbf{S}$, $\alpha$)}
    $}
    \State{$\mathbf{E_{Tess}}\gets \{\emptyset\}$}
    \For{$x_1, x_2 \in E$}
        \State{$r_1, r_2\gets  \mathbf{L}[\mathcal{N}_{x_1}] \cap \mathbf{L}[\mathcal{N}_{x_2}]$} 
        \State{\#Finding the generator points of the shared regions}
        \State{$p_1, p_2 \gets P[r_1], P[r_2]$}
        \If{$(p_1, p_2) \perp (x_1, x_2)$}
            \State{$\mathbf{E_{Tess}} \gets \mathbf{E_{Tess}} \cup (p_1, p_2)$}
        \EndIf
    \EndFor
    \State{$\mathbf{G_{Tess}} \gets (\mathbf{P, E_{Tess}})$}
    \State{\textbf{Return} $G_{Tess}$}
  \end{algorithmic}
  \label{alg:algorithmDelaunay}
\end{algorithm}

In the algorithm \ref{alg:algorithmDelaunay}, $\mathbf{L, S, P}$ are the label or region map, SNIC boundaries and the generator points sets, respectively. $(\mathbf{X, E})$ can be interpreted as the Voronoi diagram graph with regards to the regions $\mathbf{V}(p_i)| \forall 1\leq i \leq v$ \textit{as Voronoi vertices set}, while $\mathbf{E}$ is the set of Voronoi edges. Also,  $r_1, r_2$ refers to the label map of the two shared regions in a particular Voronoi edge $e$. Finally, $p_{r_1}, p_{r_2}$ represents the generator points of the regions $r_1$ and $r_2$, respectively. In the remaining of this paper, we call these regions Voronopixels. The for loop implements the creation of Tessellation edges as follows:
\begin{equation}\label{eq:tessEdgeBuilder}
    \begin{split}
        E_{Tess} &\gets\\
        \{(p_i,p_j)\} &\iff \exists (p_i, p_j)  \in \mathbf{P}, \\
        &(p_i, p_j) \perp (x_l, x_k)\in \mathbf{X} ,\\
        &\text{ region\_id}_{neighbors_{x_l}= \text{region\_id}}\{p_i, p_j\} 
    \end{split}   
\end{equation} \,

In this equation, \( (p_i, p_j) \) represents an edge in the Delaunay triangulation, where \( p_i \) is the generator point of the SNIC-defined region. Similarly, \( (x_l, x_k) \) is an edge in the Voronoi diagram generated using the Douglas-Peucker algorithm, with \( x_l \) and \( x_k \) being Voronoi vertices at boundary intersections shared by at least three regions. Notably, \( |P| = m \) and \( |V| = v \), where \( P \) and \( V \) denote the Voronoi and Delaunay triangulation vertices, respectively.

We employ SNIC to obtain superpixels \cite{15-SNIC-achanta2017superpixels}, as it offers fast convergence and memory efficiency compared to SLIC and LSC \cite{11-LinearSpectralClustering7814265}. A detailed discussion of Algorithm \ref{alg:algorithmDelaunay} follows.

From a computational complexity perspective, SNIC (line 2) is non-iterative, operating in \( O(n) \), where \( n \) is the number of pixels. The Douglas-Peucker algorithm \cite{16-DouglasPeuker} refines superpixel boundaries by converting non-straight edges into straight lines, using \( \alpha \) as a hyperparameter for adjustment. Since the number of vertices in the Voronoi diagram and Delaunay triangulation is at most \( n \), the running time depends on the number of vertices and their degrees, specifically \( O(|X|.|E|) \), where \( |X|.|E| \ll n \). The \textit{for} loop and nested statements (lines 5–12) run in \( O(|E|) \), making the overall algorithm run in \( O(n) \).

\begin{figure}
    \centering
    \includegraphics[width=0.3\textwidth]{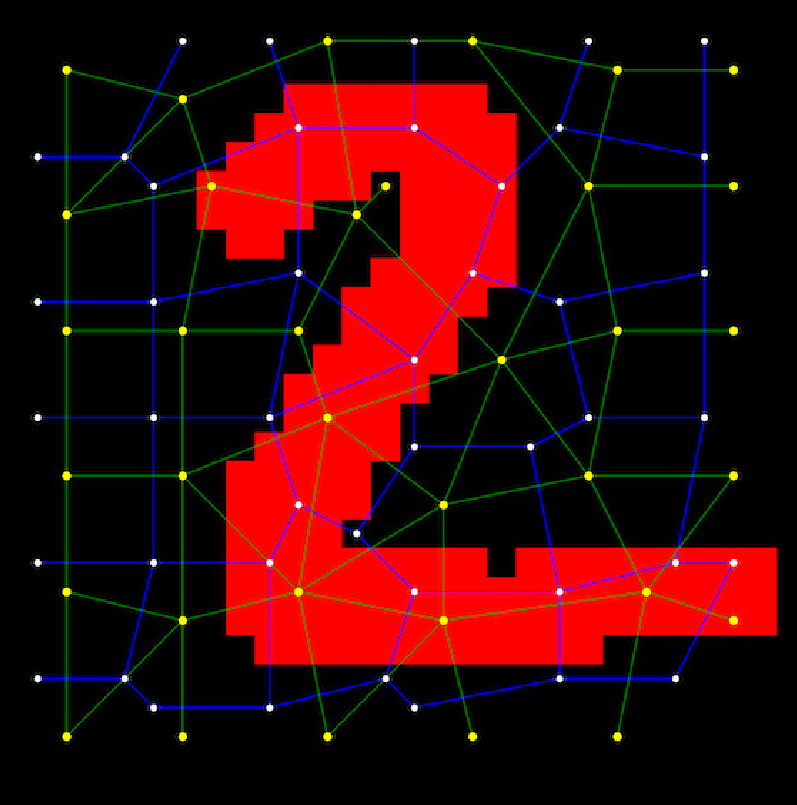}
    \caption{Voronoi diagrams (the blue graph) and the related Delaunay triangulation (the green graph) on an image of MNIST dataset created by the suggested method.}
    \label{fig:Delaunay_triangulation}
\end{figure}

By executing Algorithm \ref{alg:algorithmDelaunay}, we obtain both the Voronoi diagram—generated using the Douglas-Peucker algorithm \cite{16-DouglasPeuker}—and the corresponding Delaunay triangulation graph in \( O(n) \). As known in the field, Voronoi diagrams and Delaunay triangulations exhibit a duality relationship. An example is illustrated in Fig. \ref{fig:Delaunay_triangulation}, where the Voronoi diagram and its triangulation are represented in blue and green, respectively.

\begin{figure}
    \centering
    \includegraphics[width=0.3\textwidth]{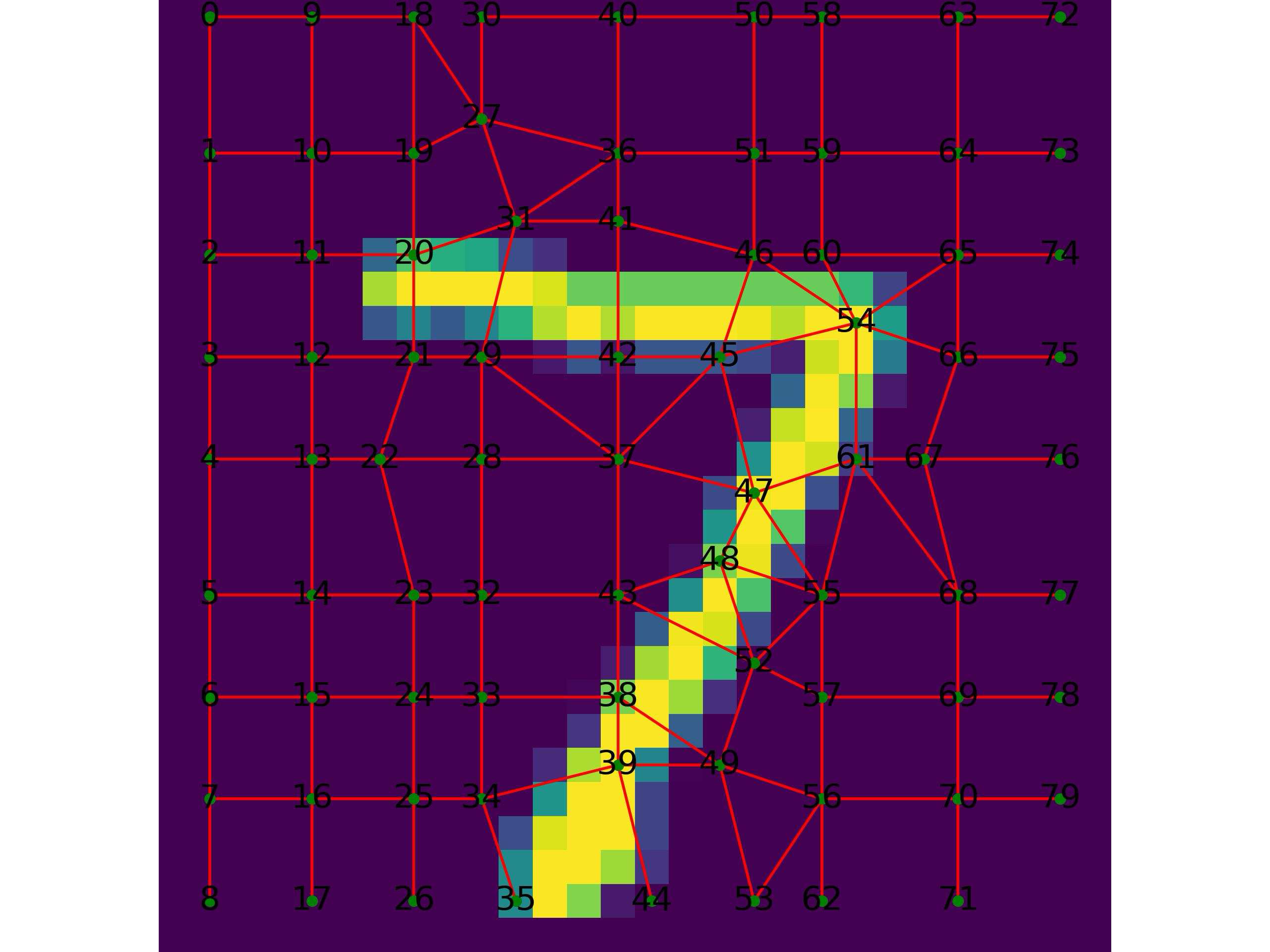}\hfill
    \includegraphics[width=0.3\textwidth]{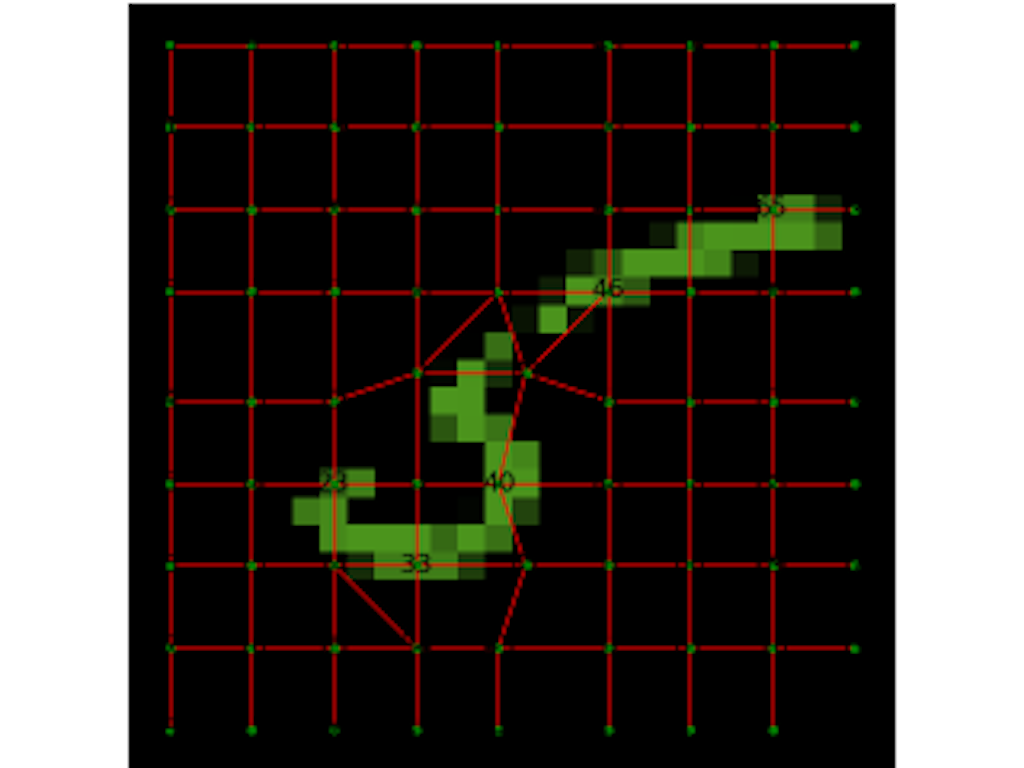}
    \includegraphics[width=0.3\textwidth]{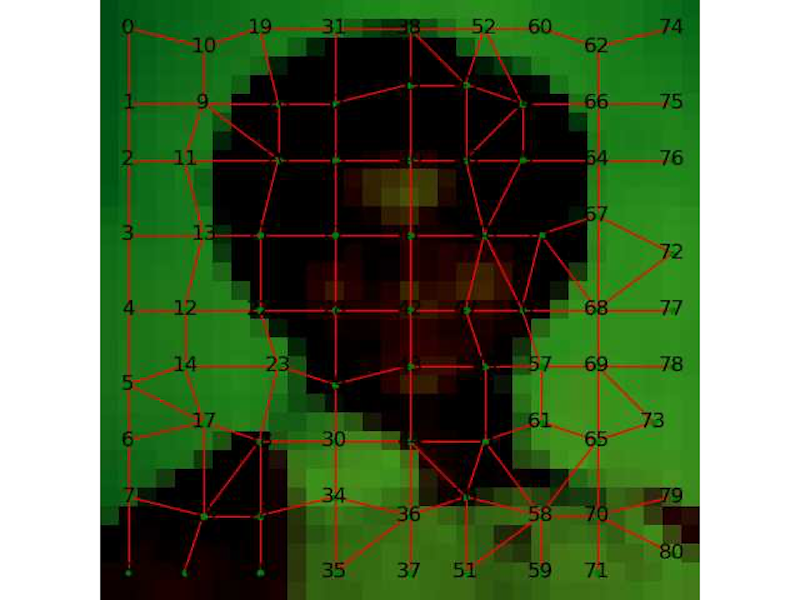}\hfill
    \caption{Delaunay triangulation obtained with approximately 81 generator points for the MNIST and CIFAR100 datasets. As shown in the image, we observe that the number of edges around the edge of the digits is deformed.}
    \label{fig:Tessellation_graph_128}
\end{figure}

Fig. \ref{fig:Tessellation_graph_128} displays the output of the proposed algorithm, highlighting increased sensitivity around the edges. The vertices near boundaries tend to have a higher number of connected edges, resulting in a more irregular topology compared to other regions. Experimentally, vertices in uniform background areas form a structure resembling a regular grid, whereas those in regions with varying color intensities exhibit greater structural deformation.

\begin{figure}[ht!]
\centering
    \begin{subfigure}[b]{0.22\textwidth}
        \includegraphics[width=\textwidth]{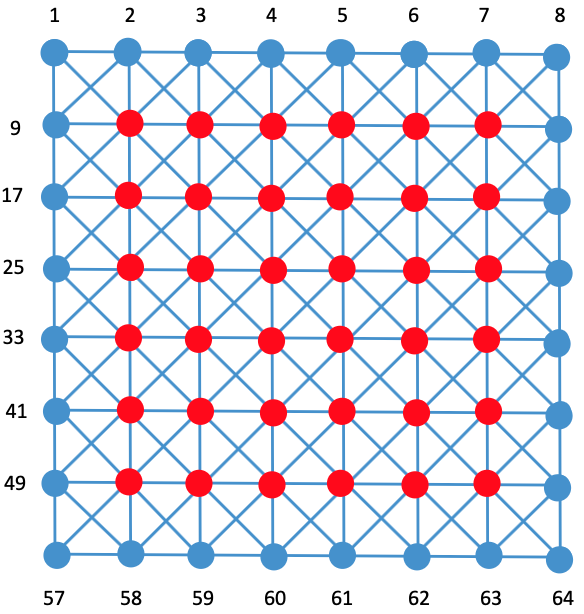}
        \caption{}
        \label{FigdensestDelaunay}
    \end{subfigure}
    \hfill
    \begin{subfigure}[b]{0.4\textwidth}
        \centering
        \includegraphics[width=\textwidth]{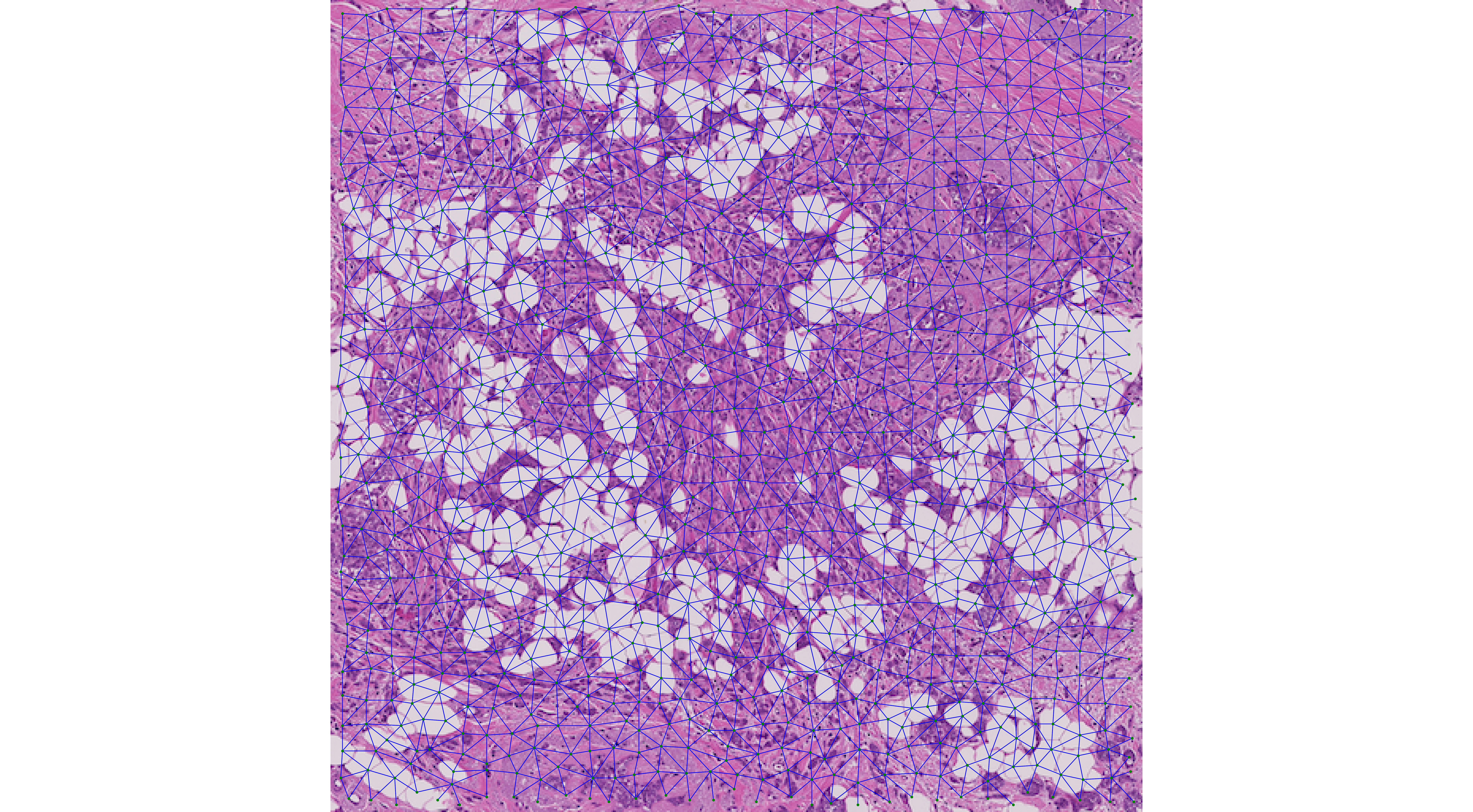}
        \caption{}
        \label{fig:regionized_TMA_img}
    \end{subfigure}
    \hfill
    \begin{subfigure}[b]{0.3\textwidth}
        \centering
        \includegraphics[width=\textwidth]{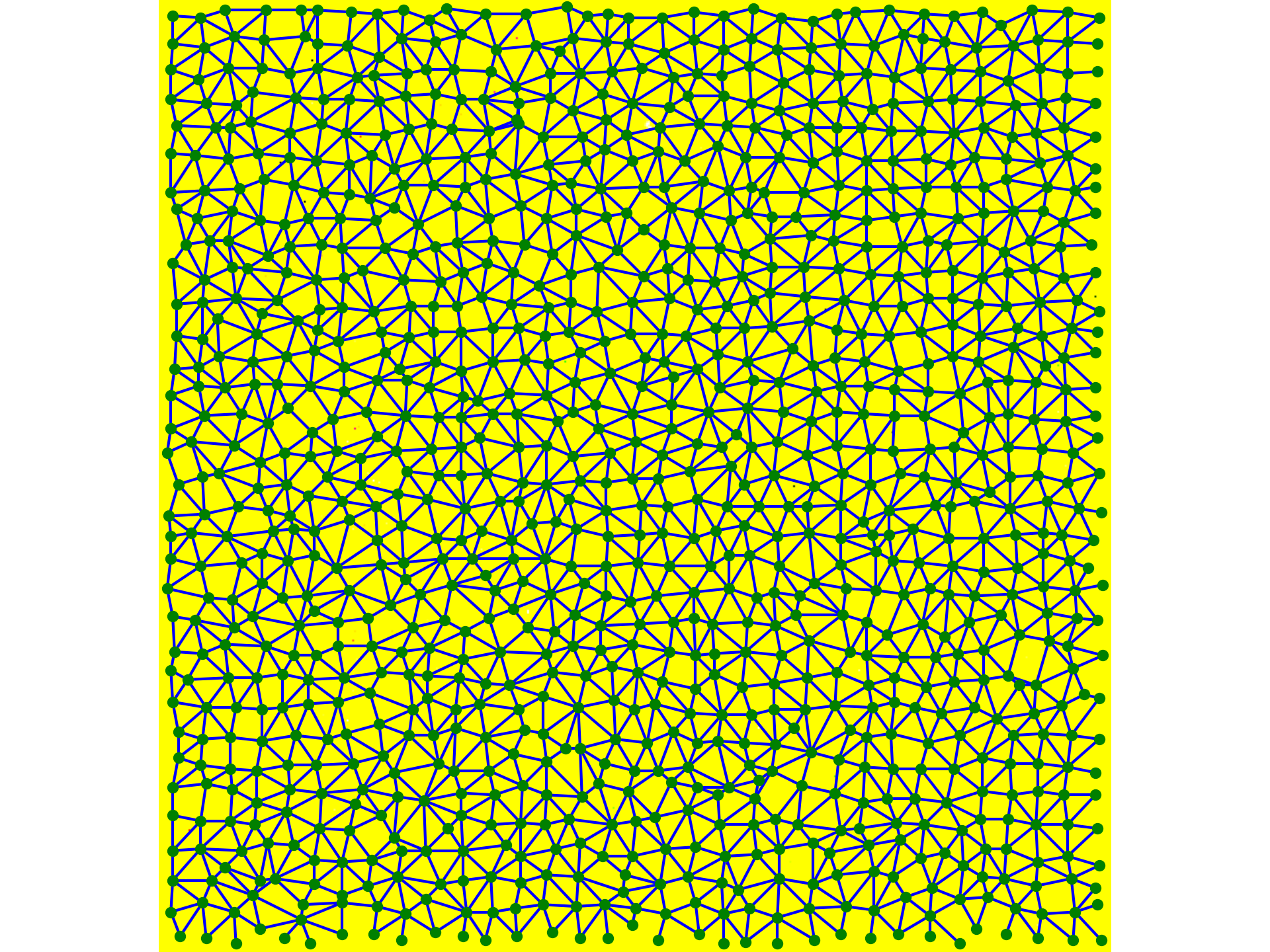}
        \caption{}
        \label{fig:voronoi_TMA_img}
    \end{subfigure}
    \caption{(a) The densest Delaunay Triangulation Graph for an image. (b) The Delaunay on an image with [7k X 7k] pixels. (c) The Delaunay Triangulation of Biomedical image in (b) without the image.}
    \label{fig:three_images}
\end{figure}

Another useful property of the graph obtained using this technique is that the densest Delaunay Triangulation graph in an image occurs when each pixel acts as a node in the graph, Fig. \ref{FigdensestDelaunay}, In this particular example, the degree of nodes 1, 8, 57, and 64 is three while all other nodes on the edge of the image are 5. Also, the highest degree in this graph is related to the node within the graph which is 8 and is shown in red. Figure \ref{fig:regionized_TMA_img} illustrates a high-resolution Tissue Microarray(TMA) image with [7000 x 7000] pixels which the Delaunay Triangulation graph is plotted on the top of it. The number of nodes in this example is about 1024 nodes. Fig. \ref{fig:voronoi_TMA_img} is the solely Delaunay triangulation graph plotted without the image for clarity.

\section{Experiments}
To evaluate the performance of our proposed Voronoi Graph Convolutional Network (VGCN) architecture, we applied extensive experiments using three standard datasets: MNIST, FashionMNIST, and CIFAR-10 \cite{CIFAR10} as shown in details in the table \ref{table:dataset_comparison}. Each dataset was transformed into graph-based representations using the methodologies described in Section 3, ensuring consistency and computational efficiency in preprocessing. We set the number of $k=64$ as the regular number of generator points for the $28\times 28$, and $k=128$ for $32 \times 32$ images. The reason why Cifar-10 has larger $k$, is that CIFAR-10 is more complicated than the other two datasets. We also set $S$ in Equation (\ref{eq:SLIC_formula}) to 50 to balance the consistency, compactness and the geolocation of the generator points. The number of epochs and the early stopping parameters set to 300 and 10 respectively to avoid from infinite training and overfitting. The learning rate in our experiments was set to 0.001, and remains fixed across all experiments. For all batches, the batch size is set to 128 graphs. The number of inputs features is set to 3 or 5 grayscale or RGB images used for creating graphs, which is the location of the generator points and the means of the normalized colour[s] intensity for each convex hull in the Delaunay triangulation. The training and test size ratio is 0.2, which means 20\% of the dataset is used as a test set. 
To test the robustness of the proposed method, Gaussian noise (μ=0, σ=0.1) and salt-and-pepper noise (noise density = 0.02) were added to the input images. These noise types simulate real-world distortions and evaluate the VGCN’s ability to generalize under varying conditions.

\begin{table}[h]
    \centering
    \begin{tabular}{|l|c|c|c|c|}
        \hline
        \textbf{Dataset} & \textbf{\# Images} & \textbf{\# Labels} & \textbf{Image size} & \textbf{Type} \\ \hline
        CIFAR-10 & 60,000 & 10 & 32x32 & RGB \\ \hline
        FashMNIST & 60,000 & 10 & 28x28 & Gray \\ \hline
        MNIST & 60,000 & 10 & 28x28 & Gray \\ \hline
    \end{tabular}%
    \caption{Comparison of CIFAR-10, Fashion-MNIST, and MNIST datasets.}
    \label{table:dataset_comparison}
\end{table}

A summary of the model used for this study is shown in Table \ref{tab:GATMultiHead}. After the last layer of the GCN we used global pooling and pass the output to the MLP layers. The input of the model is the set of Delaunay triangulation vertices and the set of edges, $G_{Tess} = (\mathbf{P}, E_{Tess})$ and the aim of the model is a classification task.

We have used Pytorch-Geometric to handle the graph and for utilizing the GAT architecture \cite{20-pygeometric-Fey2019PyTorch}. A summary of the model is shown in Table \ref{tab:GATMultiHead}.

\begin{table}[h!]
    \centering
    \resizebox{0.7\textwidth}{!}{
        \begin{tabularx}{\linewidth}{|>{\centering\arraybackslash}X|>{\centering\arraybackslash}X|>{\centering\arraybackslash}X|>{\centering\arraybackslash}X|}
        \hline
        \textbf{Layer} & \textbf{Input Features} & \textbf{Output Features} & \textbf{Details} \\ \hline
        GATConv & 3 & 32 & head=2 \\ \hline
        BatchNorm & N/A & 64 & N/A \\ \hline
        GATConv & 64 & 64 & head=2 \\ \hline
        BatchNorm & N/A & 128 & N/A \\ \hline
        GATConv & 128 & 64 & head=2 \\ \hline
        BatchNorm & N/A & 128 & N/A \\ \hline
        Linear & 128 & 32 & bias=True \\ \hline
        BatchNorm & N/A & 32 & N/A \\ \hline
        Linear & 32 & 32 & bias=True \\ \hline
        Linear & 32 & 10 & bias=True \\ \hline
        \end{tabularx}
    }
    \caption{GATMultiHead model structure.}
    \label{tab:GATMultiHead}
\end{table}

\section{Performance Evaluation and Results}

The experimental findings indicate that the proposed VGCN performs competitively across the tested datasets.
On MNIST, VGCN achieved 95.5\% accuracy, comparable to state-of-the-art methods while maintaining computational efficiency. On FashionMNIST, which contains more complex visual features, VGCN reached 81.4\%, showcasing its ability to capture relational and contextual information. For CIFAR-10, VGCN achieved 45\% accuracy, indicating areas for further optimization on higher-resolution datasets.

In the comparison, we include the following methods: Image Classification using Graph Neural Networks with Multiscale Wavelet Superpixels (Wave Mesh) \cite{Wavelet_vasudevan2023image}, Superpixel Image Classification with Graph Attention Networks (HGNN) \cite{RAG_avelar2020superpixel}, Spatial Graph Convolutional Networks (SGCN) \cite{danel2020spatial}, and Topological Graph Neural Networks (TOGL) \cite{horn2021topological}.  

Table \ref{tab:combined_gnn_results} presents the best-reported results from our implementations compared to those in the original papers, organized by dataset. To enhance readability, "SP" denotes superpixels, and the best result achieved by VGCN for each dataset is highlighted in bold, alongside the best results from the authors' implementations. Additionally, "Grid" is listed for methods incorporating specific graph structures, while "PC" or "Mac" is noted where applicable to indicate system differences.

\begin{table}[!htbp]
    \centering
    \resizebox{\textwidth}{!}{ 
        \begin{tabular}{>{\raggedright\arraybackslash}p{2.5cm} >{\centering\arraybackslash}p{2.5cm} >{\centering\arraybackslash}p{2.5cm} >{\centering\arraybackslash}p{2.5cm} >{\centering\arraybackslash}p{2.5cm} >{\centering\arraybackslash}p{2.5cm} >{\centering\arraybackslash}p{2.5cm}}
            \toprule
            \textbf{Method} & \textbf{MNIST (Authors')} & \textbf{MNIST (Ours)} & \textbf{FashionMNIST (Authors')} & \textbf{FashionMNIST (Ours)} & \textbf{CIFAR-10 (Authors')} & \textbf{CIFAR-10 (Ours)} \\ 
            \midrule
            \textbf{HGNN-2Head}     & 96.19 & \textbf{96.22}    & 83.07 & \textbf{82.95}    & 45.93 & \textbf{66.08} \\
            \textbf{SGCN (Grid)}    & \textbf{99.61} & --       & -- & --                   & -- & 49.27 \\
            \textbf{SGCN (SP, PC)}  & 95.95 & 93.07            & -- & --                   & -- & -- \\
            \textbf{SGCN (SP, Mac)} & 95.95 & 89.76            & -- & --                   & -- & -- \\
            \textbf{GCN-4}          & 90.0$\pm$0.3 & 77.25     & -- & 64.31                & 54.2$\pm$1.5 & 47.38 \\
            \textbf{GCN3-TOGL-1}    & 95.5$\pm$0.2 & 90.60     & -- & 76.95                & \textbf{61.7$\pm$1.0} & 37.41 \\
            \textbf{VGCN (ours)}    & -- & \underline{\textbf{95.5}} & -- & \underline{\textbf{81.4}} & -- & \underline{\textbf{45}} \\
            \bottomrule
        \end{tabular}
    }
    \caption{Comparison of our implementation of three GNN model performances on each dataset to the author's reported results.}
    \label{tab:combined_gnn_results}
\end{table}

To achieve a fair comparison, all accuracies are reported based on the number of vertices. We converted each image into the corresponding Delaunay triangulation with 64-85 vertices for all three datasets. Also, we did not fine-tune the model's hyper-parameters to achieve the best accuracy level.

\textbf{Computational Efficiency:} 
One of the key advantages of our VGCN model is its computational efficiency. Using Voronoi diagrams for graph construction significantly reduces the preprocessing time compared to traditional superpixel methods like SLIC. The overall time complexity for converting an image into a graph using our method is $O(n)$, where $n$ is the number of pixels in the image. This efficiency is crucial for applications requiring real-time image processing and classification. Also, in our work, the speed of the GCN is much faster than that of traditional GCNs. This is because of the sparsity of the adjacency matrix for the Delaunay triangulation graph. The average number of edges for each vertex is approximately $6.25$, indicating that the graph is sparse compared to all other methods.

\textbf{Robustness}:
To test the robustness of the proposed method, Gaussian noise ($\mu=0$, $\sigma=0.12)$ and salt-and-pepper noise (noise density = $0.02$) were added to the input images. These noise types simulate real-world distortions and evaluate the VGCN’s ability to generalize under varying conditions. Under the influence of Gaussian and salt-and-pepper noise, the VGCN maintained a high level of accuracy, particularly on MNIST and FashionMNIST datasets. The graph-based representation’s resilience to noise can be attributed to its ability to capture both local and global features effectively.

In terms of preprocessing, to convert an image to the corresponding graph, we tracked the time for each image in a single and multi-processing manner. Converting an image into a graph just took 0.023 seconds. We used 12 core CPUs to convert the MNIST and FashionMNIST images to graphs, and it took 110 seconds to convert 60,000 images. The baseline model was a bit faster, around 10 seconds. The reason behind that is that converting the image into a Delaunay triangulation needed a few more computations than SNIC. On the other hand, the VGCN algorithm took less time to converge.

Finally, the results of our experiments highlight several key points regarding the performance and applicability of the VGCN model and its efficency in graph construction. Using Voronoi diagrams for image partitioning and graph construction efficiently and effectively captures the relational context of image regions. This efficiency makes the VGCN model suitable for real-time applications. The number of regions/superpixels is shown in table 4. As it shown in the table, we found that when an image is more complex, the number of regions, nodes in the graphs, should be larger.

\begin{table}[h]
    \label{tblVertices_comparison}
    \centering
    \resizebox{.7\textwidth}{!}{
        \begin{tabularx}{\linewidth}{|>{\centering\arraybackslash}X|>{\centering\arraybackslash}X|>{\centering\arraybackslash}X|>{\centering\arraybackslash}X|}
            \hline
            \textbf{Model} & \textbf{MNIST vertices} & \textbf{Fashion MNIST vertices} & \textbf{CIFAR-10 vertices} \\ \hline
        VGCN & 64 & 64 & 150 \\ \hline
        Superpixel GAT & 75 &75 & 150 \\ \hline
        \end{tabularx}
        }
    \caption{Number of vertices used by different models on various datasets.}
\end{table}

In addition to these comparisons, we have used three different machine learning architectures, GCN, GAT, and VGCN for tracking the times spent for each epoch. We changed the first architecture as VGCN, Voronoi Graph Convolution Network which is used for both GCN and GAT architectures. We used the architecture used for this research to analyze the VGCN  and GAT in terms of speed, table \ref{tab:time_comp_NVGCN}. We used the same setting for the model for VGCN as the table \ref{tab:GATMultiHead} for fairness.

\begin{table}[h]
    \centering
	\begin{tikzpicture}
		\begin{axis}[
            width=0.6\textwidth, 
            height=0.3\textwidth, 
			xlabel={Epoch},
			ylabel={Time (minutes)},
			legend style={
                font=\tiny, 
                at={(0.5,-0.2)}, 
                anchor=north,
                legend columns=-1, 
            },
            legend image post style={scale=0.7}, 
			grid=both,
			ymin=0, ymax=0.6,
			xtick={1,2,3,4,5,6,7,8,9,10},
			ytick={0,0.1,0.18,0.2,0.3,0.4,0.5,0.6},
			tick label style={font=\scriptsize}, 
			label style={font=\small} 
			]
			
			\addplot[color=blue,mark=*] coordinates {
				(1, 0.19) (2, 0.19) (3, 0.20) (4, 0.20) (5, 0.21)
				(6, 0.21) (7, 0.21) (8, 0.20) (9, 0.20) (10, 0.20)
			};
			\addlegendentry{Regular GCN}
			
			\addplot[color=red,mark=square*] coordinates {
				(1, 0.16) (2, 0.18) (3, 0.17) (4, 0.18) (5, 0.18)
				(6, 0.18) (7, 0.18) (8, 0.18) (9, 0.18) (10, 0.18)
			};
			\addlegendentry{VGCN\_pyg}
			
			\addplot[color=green,mark=triangle*] coordinates {
				(1, 0.51) (2, 0.50) (3, 0.53) (4, 0.51) (5, 0.50)
				(6, 0.51) (7, 0.50) (8, 0.51) (9, 0.51) (10, 0.52)
			};
			\addlegendentry{Regular GAT (2 heads)}
			
			\addplot[color=orange,mark=diamond*] coordinates {
				(1, 0.30) (2, 0.31) (3, 0.31) (4, 0.30) (5, 0.31)
				(6, 0.30) (7, 0.31) (8, 0.29) (9, 0.29) (10, 0.29)
			};
			\addlegendentry{Regular GAT (1 head)}
			
		\end{axis}
    \end{tikzpicture}
    \caption{Time comparison for three different models compared to the VGCN. As the plot shows, the VGCN has spent the least time compared to other methods.}
    \label{tab:time_comp_NVGCN}
\end{table}

\section{Conclusion and Future Work}
This paper demonstrates that integrating Graph Neural Networks with Delaunay triangulations advances image classification. Despite challenges, particularly with complex datasets, this approach exhibits computational efficiency and robustness on standard benchmarks. The combination of GCNs and Delaunay triangulation graphs enhances classification accuracy and effectiveness while opening new research directions in computer vision and image processing.
While integrating GCNs with advanced techniques is promising, challenges remain, especially in managing the computational complexity of large-scale graph representations, even in fields such as recommendation systems and virtual reality. 

Future work could focus on optimizing graph construction and feature extraction algorithms, as well as exploring the adaptability of these methods for real-time processing and classification of dynamic images.
Additionally, this research can be extended in other ways, including experimenting with different metrics or distance functions for constructing Delaunay triangulation graphs.
\\

\noindent \textbf{Acknowledgments:}
This research work has been partially supported by the Natural Sciences and Engineering Research Council of Canada, NSERC, and the Faculty of Science, University of Windsor.

\printbibliography[heading=subbibintoc]

\end{document}